\documentclass[11pt,a4paper]{article}
\usepackage[hyperref]{acl2020}
\usepackage{times}
\usepackage{latexsym}
\usepackage{xspace,mfirstuc,tabulary}
\usepackage{amsmath}
\usepackage{multirow}
\usepackage{comment}
\usepackage{xcolor}
\usepackage{ulem}
\usepackage{paralist}

\usepackage[utf8]{inputenc}
\usepackage{pgfplots}
\pgfplotsset{compat=newest}
\usepgfplotslibrary{groupplots}
\usepgfplotslibrary{dateplot}

\makeatletter
\newcommand{\@emptybiblabel}[1]{}
\makeatother
\usepackage{color}

\definecolor{myblue}{rgb}{0, 57, 230}
\definecolor{airforceblue}{rgb}{0.36, 0.54, 0.66}
\definecolor{navyblue}{rgb}{0, 0, 128}
\definecolor{ceruleanblue}{rgb}{0.16, 0.32, 0.75}
\definecolor{cornflowerblue}{rgb}{0.39, 0.58, 0.93}
\definecolor{denim}{rgb}{0.08, 0.38, 0.74}

\definecolor{azure(colorwheel)}{rgb}{0.0, 0.5, 1.0}
\definecolor{cornellred}{rgb}{0.7, 0.11, 0.11}

\definecolor{lemon}{rgb}{1.0, 0.97, 0.0}
\definecolor{amber(sae/ece)}{rgb}{1.0, 0.49, 0.0}
\definecolor{cadmiumorange}{rgb}{0.93, 0.53, 0.18}
\definecolor{darkorange}{rgb}{1.0, 0.55, 0.0}
\definecolor{debianred}{rgb}{0.84, 0.04, 0.33}
\definecolor{deepcarmine}{rgb}{0.66, 0.13, 0.24}
\definecolor{deepcarminepink}{rgb}{0.94, 0.19, 0.22}

\definecolor{brightgreen}{rgb}{0.4, 1.0, 0.0}
\definecolor{caribbeangreen}{rgb}{0.0, 0.8, 0.6}
\definecolor{chartreuse(web)}{rgb}{0.5, 1.0, 0.0}
\definecolor{darkpastelgreen}{rgb}{0.01, 0.75, 0.24}
\definecolor{electricgreen}{rgb}{0.0, 1.0, 0.0}
\definecolor{emerald}{rgb}{0.31, 0.78, 0.47}
\definecolor{cadmiumgreen}{rgb}{0.0, 0.42, 0.24}

\definecolor{darkmagenta}{rgb}{0.55, 0.0, 0.55}
\definecolor{darklavender}{rgb}{0.45, 0.31, 0.59}

\usepackage{helvet}  
\usepackage{courier}  
\usepackage{graphicx}  
\usepackage{amssymb}
\usepackage{amsfonts}
\usepackage{paralist}
\usepackage{bbold}
\usepackage[toc,page]{appendix} 
\usepackage{float}
\usepackage{makecell}
\usepackage{subfig}

\DeclareSymbolFontAlphabet{\amsmathbb}{AMSb}%

\newcommand{\R}[0]{\amsmathbb R}

\DeclareMathOperator*{\maxp}{max-pool}

\usepackage{pict2e,picture}
\makeatletter
\DeclareRobustCommand{\Arrow}[1][]{%
\check@mathfonts
\if\relax\detokenize{#1}\relax
\settowidth{\dimen@}{$\m@th\rightarrow$}%
\else
\setlength{\dimen@}{#1}%
\fi
\sbox\z@{\usefont{U}{lasy}{m}{n}\symbol{41}}%
\begin{picture}(\dimen@,\ht\z@)
\roundcap
\put(\dimexpr\dimen@-.7\wd\z@,0){\usebox\z@}
\put(0,\fontdimen22\textfont2){\line(1,0){\dimen@}}
\end{picture}%
}
\makeatother

\aclfinalcopy 

\title{Query Focused Multi-Document Summarization with Distant Supervision}

 \author{Yumo Xu \and Mirella Lapata\\
 Institute for Language, Cognition and Computation\\
 School of Informatics, University of Edinburgh\\
 10 Crichton Street, Edinburgh EH8 9AB\\
 \texttt{yumo.xu@ed.ac.uk}, 
 \texttt{mlap@inf.ed.ac.uk}}

\date{}

\makeatletter
\newcommand{\thickhline}{%
    \noalign {\ifnum 0=`}\fi \hrule height 1pt
    \futurelet \reserved@a \@xhline
}
\makeatother

\begin{document}
\maketitle
\begin{abstract}
 
  We consider the problem of better modeling query-cluster
  interactions to facilitate query focused multi-document
  summarization (QFS).
  Due to the lack of training data, existing work relies heavily on
  retrieval-style methods for estimating the relevance between queries
  and text segments. In this work, we leverage distant supervision
  from question answering where various resources are available to
  more explicitly capture the relationship between queries and
  documents.  We propose a coarse-to-fine modeling framework which
  introduces separate modules for estimating whether segments are
  relevant to the query, likely to contain an answer, and
  central. Under this framework, a trained \emph{evidence} estimator
  further discerns which retrieved segments might answer the query for
  final selection in the summary. We demonstrate that
our framework outperforms strong comparison systems on standard QFS
benchmarks.
\end{abstract}

\section{Introduction}
\label{sec:introduction}

Query Focused Multi-Document Summarization (QFS; \citealt{Dang:2006})
aims to create a short summary from a set of documents that answers a
specific query. It has various applications in personalized
information retrieval and recommendation engines where search results
can be tailored to an information need (e.g.,~a user might be looking
for an overview summary or a more detailed one which would allow them
to answer a specific question).

Neural approaches have become increasingly popular in single-document
text summarization
\cite{K16-1028,paulus2018a,li2017salience,P17-1099,N18-1158,D18-1443},
thanks to the representational power afforded by deeper architectures
and the availability of large-scale datasets containing hundreds of
thousands of document-summary pairs
\cite{nytcorpus,hermann-nips15,newsroom-naacl18}. Unfortunately, such
datasets do not exist in QFS, and one might argue it is unrealistic
they will ever be created for millions of queries, across different
domains (e.g.,~news vs user reviews), and languages. In addition to
the difficulties in obtaining training data,  another
obstacle to the application of end-to-end neural models is the size
and number of source documents which can be very large. It is
practically unfeasible (given memory limitations of current hardware)
to train a model which encodes all of them into vectors and
subsequently generates a summary from them.

In this paper we propose a coarse-to-fine modeling framework for
extractive QFS which incorporates a \emph{relevance} estimator for
retrieving textual segments (e.g., sentences or longer passages)
associated with a query, an \emph{evidence} estimator which further
isolates segments likely to contain answers to the query, and a
\emph{centrality} estimator which finally selects which segments to
include in the summary. The vast majority of previous work \cite{wan2007manifold,wan2008using,wan2009graph,wan2014ctsum}
creates summaries by ranking textual segments (usually
sentences) according to their relationship (e.g., similarity) to other
segments \emph{and} their relevance to the query. In other words,
relevance and evidence estimation are subservient to estimating the
centrality of a segment (e.g., with a graph-based model). We argue
that disentangling these subtasks allows us to better model the query
and specialize the summaries to specific questions or topics
\cite{katragadda2009query}. A coarse-to-fine approach is also
expedient from a computational perspective; at each step the model
processes a decreasing number of segments (rather than entire
documents), and as a result is insensitive to the original input size
and more scalable.

Our key insight is to treat evidence estimation as a question
answering task where a cluster of potentially relevant documents
provides support for answering a query
\cite{baumel2016topic}. Advantageously, we are able to train the
evidence estimator on existing large-scale question answering datasets
\cite{rajpurkar2016squad,joshi2017triviaqa,yang2018hotpotqa},
alleviating the data paucity problem in QFS. Existing QFS systems \cite{wan2007manifold,wan2008using,wan2009graph,wan2014ctsum}
employ classic retrieval techniques (such as TF-IDF)
to estimate the affinity between query-sentence pairs. Such techniques
can handle short keyword queries, but are less appropriate in QFS
settings where query narratives can be long and complex.  We argue
that a trained evidence estimator might be better at performing
\textit{semantic matching} \cite{guo2016deep} between queries and
document segments. To this effect, we experiment with two popular QA
settings, namely answer sentence selection
\cite{heilman2010tree,yang2015wikiqa} and machine reading
comprehension \cite{rajpurkar2016squad} which operates over passages
than isolated sentences. In both cases, our evidence estimators take
advantage of powerful pre-trained encoders such as BERT
\cite{devlin2019bert}, to better capture semantic interactions between
queries and text units.

Our contributions in this work are threefold: we propose a
coarse-to-fine model for QFS which we argue allows to introduce
trainable components taking advantage of existing datasets and
pre-trained models; we capitalize on the connections of QFS with
question answering and propose different ways to effectively estimate
the query-segment relationship; we provide experimental results on two
benchmark datasets (DUC 2006, 2007) which show that our model
consistently outperforms strong comparison systems in both automatic
and human evaluation.

\section{Related Work}
\label{sec:related-work}

\begin{figure*}[t]
  \centering
  \includegraphics[width=16cm]{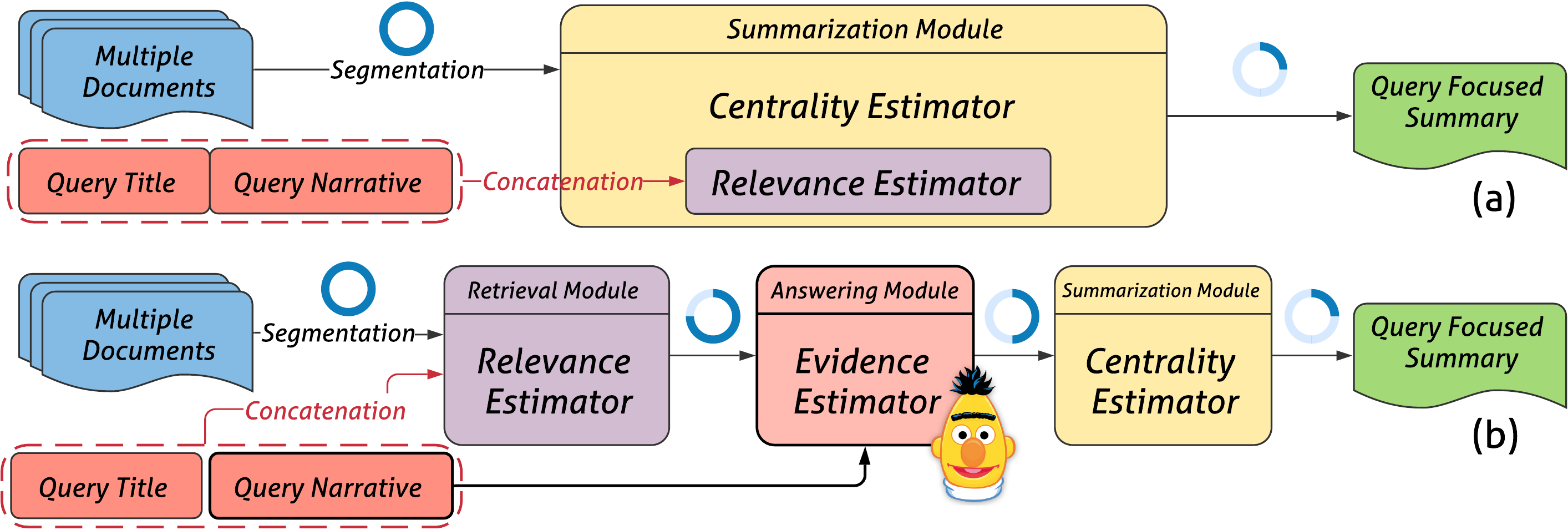}
  \caption{\label{fig:framework} Classic (a) and proposed framework
    (b) for query-focused summarization.  The classic approach
    involves a relevance estimator nested within a summarization module
    while our framework takes document clusters as input, and
    \emph{sequentially} processes them with three individual modules
    (relevance, evidence, and centrality estimators). The blue circles
    indicate a coarse-to-fine estimation process from original
    articles to final summaries where modules gradually discard
    segments (i.e., sentences or passages). With regard to evidence
    estimation, we adopt pretrained BERT \citep{devlin2019bert} which
    is further fine-tuned with distant signals from question
    answering.}
\end{figure*}
\label{sec:relwork}


Existing research on query-focused multi-document summarization
largely lies on extractive approaches, where systems usually take as
input a set of documents and select the sentences most relevant to the
query for inclusion in the summary. 

Centrality-based approaches have generally shown strong performance in
QFS.  In Figure~\ref{fig:framework}(a), we provide a sketch of classic
centrality-based approaches where all sentences within a document
cluster, together with their query relevance, are jointly considered
in estimating centrality.  Under this framework,
\newcite{wan2008using} propose a topic-sensitive version of the Markov
Random Walk model which integrates sentence relevance, while
\newcite{wan2014ctsum} incorporate predictions about information
certainty.  Another line of graph-based work uses manifold-ranking
algorithms \cite{wan2007manifold,wan2009graph,wan2009topic} to
estimate sentence importance scores based on the assumption that
nearby points are likely to have similar rankings.  To alleviate the
mismatch between queries and document sentences,
\newcite{nastase2008topic} employs Wikipedia as a knowledge resource
for query expansion.

More recently, \newcite{li2015reader} estimate the salience of text
units within a sparse-coding framework by additionally taking into
account reader comments (associated with news reports).
\citet{li2017cascaded} use a cascaded neural attention model to find
salient sentences, whereas in follow-on work \citet{li2017salience}
employ a generative model together with a data reconstruction
model. The generative model maps sentences to a latent semantic space
based on variational autoencoders
\cite{kingma2013auto,rezende2014stochastic} while the reconstruction
model estimates sentence salience.  There are also feature-based
approaches achieving good results by optimizing sentence selection
under a summary length constraint \cite{feigenblat2017unsupervised}.

In contrast to previous work, our proposed framework does not
simultaneously perform segment selection and query matching. We
introduce a coarse-to-fine approach that incorporates progressively
more accurate components for selecting segments to include in the
summary, making model performance relatively insensitive to the number
and size of input documents. Drawing inspiration from recent work on
QA, we take advantage of existing datasets in order to reliably
estimate the relationship between the query and candidate segments. We
focus on two QA subtasks which have attracted considerable attention
in the literature, namely \textit{answer sentence selection} which
aims to extract answers from a set of pre-selected sentences
\cite{heilman2010tree,yao2013answer,yang2015wikiqa}
and \textit{machine reading comprehension}
\cite{rajpurkar2016squad,welbl2018constructing,yang2018hotpotqa},
which aims at answering a question after processing a short text
passage \cite{chen2018neural}.
An outstanding difference between QA and QFS is that extractive QA
models aim at finding the \emph{best} answer in a span or sentence,
while QFS models learn to extract a \emph{set} of sentences
considering user preferences and the content of the input
documents \cite{wan2008using,wan2014ctsum}.

\section{Problem Formulation}
\label{sec:problem-formulation}

Let $\mathcal Q$ denote an information request and
\mbox{$\mathcal{D} = \{d_1, d_2, \dots, d_M\}$} a set of topic-related
documents.  It is often assumed (e.g.,~in DUC competitions) that
$\mathcal Q$~consists of a short title (e.g., \textsl{Amnesty
  International}) highlighting the topic of interest, and a query
narrative which is considerably longer and detailed (e.g.,
\textsl{What is the scope of operations of Amnesty International and
  what are the international reactions to its activities?}). 

We illustrate our proposed framework in Figure~\ref{fig:framework}(b).
We first decompose documents into segments, i.e.,~passages or
sentences, and retrieve those which are most relevant to
query~$\mathcal{Q}$ (Relevance Estimator).  Then, a trained estimator
quantifies the semantic match between selected segments and the query
(Evidence Estimator) to further isolate segments for consideration in
the output summary (Centrality Estimator). We propose two variants of
our evidence estimator; a context agnostic variant infers evidence
scores over individual sentences, while a context aware one infers
evidence scores for tokens within a passage which are further
aggregated into sentence-level evidence. Passages might allow for
semantic relations to be estimated more reliably since neighboring
context is also taken into account.

\subsection{Relevance Estimator}
\label{sec:relevance-estimator}


\paragraph{Document Segmentation}
Our QFS system operates over documents within a cluster which we
segment into sentences.  The latter serve as input to the context
agnostic evidence estimator. For the context aware variant, we obtain
passages with a sliding window over continuous sentences in the same
document. Considering the maximum input length BERT allows (512
tokens) and the query length (to be later concatenated with passages),
we set the maximum passage size to 8~sentences (with maximum sentence
length of 50~tokens).  To ensure all sentences are properly
contextualized, we use a stride size of 4 sentences to create
overlapping passages.

\paragraph{Adaptive Retrieval} 

During inference, we first retrieve the top $k^{\text{IR}}$ answer
candidates (i.e.,~sentences or passages) which are subsequently
processed by our evidence estimator.  We do this following an adaptive
method that allows for a variable number of segments to be selected
for each query.  Specifically, for the $i$th query-cluster pair, we
first rank all segments in the cluster based on term frequency with
respect to the query, and determine~$k^{\text{IR}}_i$ such that it
reaches a fixed threshold~$\theta \in [0, 1]$.
Formally,~$k^{\text{IR}}_i$, the number of retrieved segments, is
given by:
\begin{equation}
	k^{\text{IR}}_i = \max_k \sum^k_{j=1} r_{i, j} < \theta
\end{equation}
where $r_{i, j}$ is the relevance score for segment~$j$ (normalized
over segments in the $i$th cluster). 
Although we adopt term frequency as our relevance estimator, there is
nothing in our framework which precludes the use of more sophisticated
retrieval methods \cite{dai2019deeper,yilmaz2019cross}. We
investigated approaches based on term frequency-inverse sentence
frequency \cite{allan2003retrieval} and BM25 \cite{robertson2009probabilistic}, however, we
empirically found that they are inferior, having a bias towards
shorter segments which are potentially less informative for
summarization.

\subsection{Evidence Estimator}
\label{sec:evidence-estimator}
We argue that relevance matching is not sufficient to capture the
semantics expressed in the query narrative and its relationship to the
documents in the cluster. We therefore leverage distant supervision
signals from existing QA datasets to train our evidence estimator and
use the trained estimators to rerank answer candidates selected from
the retrieval module. For the $i$th cluster, we select the top
$\min \{k^{\text{QA}}, k_i^{\text{IR}} \}$ candidates as answer
evidence (where~$k^{\text{QA}}$ is tuned on the development set).

\paragraph{Sentence Selection} 

Let~$\mathcal Q$ denote a query (in practice a sequence of tokens) and
$\{\mathcal S_1, \mathcal S_2, \dots, \mathcal S_N\}$ the set of
candidate answers (also token sequences) obtained from the retrieval
module. Our learning objective is to find the correct answer(s) within
this set.  We concatenate query~$\mathcal Q$ and candidate
sentence~$\mathcal S$ into a sequence \texttt{[CLS]}, $\mathcal Q$,
\texttt{[SEP]}, $\mathcal S$, \texttt{[SEP]} to serve as input to a BERT encoder
(we pad each sequence in a minibatch of~$L$ tokens).
The \texttt{[CLS]} vector serves as input to a single layer
neural network to obtain the distribution over positive and negative
classes: 
\begin{equation}
\hspace*{-.2cm}p^{(i)}_0 = \frac{1}{Z} \exp\left({t}_i^\intercal {W}_{:,0}\right), 
	p^{(i)}_1 = \frac{1}{Z} \exp\left({t}_i^\intercal {W}_{:,1}\right)
\end{equation}
where $Z = \sum_c \exp\left({{t}_i ^T {W}_{:,c}}\right)$ and matrix
${W} \in \R^{d\times2}$ is a learnable parameter.  We use a cross
entropy loss where 1~denotes that a sentence contains the answer (and
0~otherwise):
\begin{equation}
	\mathcal{L} = -\sum_{i=1}^N (y\log p^{(i)}_1 + (1 - y)\log p^{(i)}_0).
\end{equation}

We treat the probability of the positive class as evidence score
$q=p^{(i)}_1\in (0, 1)$ and use it to rank all retrieved segments for
each query.

\paragraph{Span Selection} 

A span selection model allows us to capture more faithfully the
answer, its local context and their interactions.  Again,
let~$\mathcal Q$ denote a query token sequence and~$\mathcal P$ a
passage token sequence.  Our training objective is to find the correct
answer span in $\mathcal P$.  Similar to sentence selection, we
concatenate the query $\mathcal Q$ and the passage $\mathcal P$ into a
sequence \texttt{[CLS]}, $\mathcal Q$, \texttt{[SEP]}, $\mathcal P$,
\texttt{[SEP]} and pad it to serve as input to a BERT encoder.  Let
${T} = [{t}_i]_{i=1}^N$ denote the contextualized vector
representation of the entire sequence obtained from BERT. We feed~$T$
into two separate dense layers to predict probabilities
$p_{\mathcal{S}}$ and $p_{\mathcal{E}}$:
\begin{gather}
	p^{(i)}_{\mathcal{S}} = 
	\frac{\exp\left({t}_i^\intercal {w}_{\mathcal{S}}\right)}
	{\sum_j 
	\exp\left({t}_j ^\intercal {w}_{\mathcal{S}}\right)}\\
	p^{(i)}_{\mathcal{E}} = 
	\frac{\exp\left({t}_i^\intercal {w}_{\mathcal{E}}\right)}
	{\sum_j 
	\exp\left({t}_j ^\intercal {w}_{\mathcal{E}}\right)}
\end{gather} 
where ${w}_{\mathcal{S}}$ and ${w}_{\mathcal{E}}$ are two learnable
vectors denoting the beginning and end of the (answer) span,
respectively.  During training we optimize the log-likelihood of the
correct start and end positions.  For passages without any correct
answers, we set these to~0 and default to the \texttt{[CLS]} position.

At inference time, to allow comparison of results across passages, we
remove the final softmax layer over different answer spans.
Specifically, we first calculate the (unnormalized) start and end
scores for all tokens in a sequence:
\begin{equation}
  {u} = \exp\left({T} {w}_\mathcal{S}\right), 
  {v} = \exp\left({T} {w}_\mathcal{E}\right).
\end{equation}
And collect sentence scores from token scores as follows. For
each sentence starting at token~$i$ and ending at token~$j$, we obtain
score matrix ${Q}$ via:
\begin{gather}
	\tilde{{Q}} = \left( {u}_{[i:j]} {v}_{[i:j]}^
	\intercal {A} \right)^{\frac{1}{2}} \\
	{Q} = \tanh(\tilde{{Q}})
\end{gather}
where ${A}$~is an upper triangular matrix masking all illegitimate
spans whose end comes before the start.  The $\tanh$~function scales
the magnitude of extreme scores (e.g.,~scores over 100 or under 0.01),
as a means of reducing the variance of~$\tilde{{Q}}$.  We collect all
possible span scores within a sentence in matrix ${S}$ where ${S}_{i',
  j'}$ denotes the span score from token $i'$ to token $j'$ ($i \leq
i' < j' \leq j $). And finally, we use max pooling to obtain a scalar
evidence score $q$: 
\begin{equation}
	q = \maxp({Q})  \in (0, 1).
\end{equation}
It is possible to produce multiple evidence scores for the same
sentence since we use overlapping passages; we select the score with
the highest value in this case.

\subsection{Centrality Estimator}

\paragraph{Graph Construction}
Inspired by \newcite{wan2008using}, we introduce as our centrality
estimator an extension of the well-known \textsc{LexRank} algorithm
\cite{erkan2004lexrank}, which we modify to incorporate the evidence
estimator introduced in the previous section. 

For each document cluster, \textsc{LexRank} builds a graph $G=\left(
  \mathcal V, \mathcal E \right)$ with nodes~$\mathcal V$
corresponding to sentences and (undirected) edges~$\mathcal E$ whose
weights are computed based on similarity.  Specifically, matrix ${E}$
represents edge weights where each element ${E}_{i, j}$ corresponds to
the transition probability from vertex~$i$ to vertex~$j$.  The
original \textsc{LexRank} algorithm uses TF-IDF (Term Frequency
Inverse Document Frequency) to measure similarity; since our framework
operates over sentences rather than ``documents'', we use TF-ISF (Term
Frequency Inverse Sentence Frequency), with ISF defined as:
\begin{equation}
\label{eq:isf}
	\text{ISF}(w) = 1 + \log (|C| / \text{SF}(w))
\end{equation}
where $C$ is the total number of sentences in the cluster, and
$\text{SF}(w)$ the number of sentences in which~$w$ occurs. 

We integrate our evidence estimator into the original transition matrix as:
\begin{equation}
\label{eq:query}
	\tilde{E} = \phi * \tilde{q} + (1-\phi) * E 
\end{equation} 
where $\phi \in (0, 1)$ controls the extent to which query-specific
information influences sentence selection for the summarization task;
and $\tilde{q}$~is a distributional evidence vector which we obtain
after normalizing the evidence scores~${q} \in \R^{1\times |V|}$
obtained from the previous module (\mbox{$\tilde{{q}}= {q} /
  \sum_v^{|V|} {q}_v$}). 

\paragraph{Summary Generation}
In order to decide which sentences to include in the summary, a node’s
centrality is measured using a graph-based ranking algorithm
\cite{erkan2004lexrank}.  Specifically, we run a Markov chain
with~$\tilde{{E}}$ on $G$ until it converges to stationary
distribution ${e}^\ast$ where each element denotes the salience of a
sentence.  In the proposed algorithm, ${e}^\ast$ jointly expresses the
importance of a sentence in the document \emph{and} its semantic
relation to the query as modulated the evidence estimator and
controlled by~$\phi$.
We rank sentences according to ${e}^\ast$ and select the
top~$k^{\text{Sum}}$ ones, subject to a budget (e.g.,~250 words). To
reduce redundancy, we apply the diversity algorithm proposed in
\newcite{wan2008using} which penalizes the salience of sentences
according  to their overlap with those already selected to appear in
the summary.

\begin{table}[t]
\centering
\begin{tabular}{l|cccc}
  \hline
DUC-Year &  2005 &2006  & 2007\\
  \hline
\#Clusters & 	50	&    50	  &   45\\
\#Documens/Cluster 
& 32 &	   25	 &    25 \\
\#Summaries &	4-9	    & 4	  &   4\\
\#Words/Summary & 250 & 250 & 250\\
    \hline
\end{tabular}
\caption{\label{tab:duc_stats}DUC statistics.}
\end{table}

\begin{table}[t]
\tabcolsep=0.11cm
\centering
\begin{tabular}{l|rrr|rr}
  \hline
\multirow{2}*{\textbf{Dataset}} 
&\multicolumn{3}{c|}{\textbf{Sentence}}
&  \textbf{Span} \\
~  &  WikiQA & TrecQA &Total & SQuAD\\
  \hline
 \#Train & 8,672 & 53,417 & 62,089  &  130,318\\
 \#Dev  & 1,130 & 1,148 & 2,278 & 11,872\\
\hline
\end{tabular}
\caption{\label{tab:qa_stats}Question answering dataset statistics. We
  use the union of WikiQA and TrecQA for answer sentence selection and
  SQuAD for span selection.} 
\end{table} 

\section{Experimental Setup}

\paragraph{Datasets}
We performed QFS experiments on the DUC 2005-2007 benchmark datasets.
We show summary statistics in Table~\ref{tab:duc_stats}.  We used
DUC~2005 as a development set to optimize hyperparameters and
evaluated model performance on DUC 2006 and 2007 (test sets).

We used three datasets for training our evidence estimator, including
WikiQA \cite{yang2015wikiqa}, TrecQA \cite{yao2013answer}, and SQuAD
2.0 \cite{rajpurkar2018know}.
WikiQA and TrecQA are benchmarks for answer sentence selection while
SQuAD 2.0 is a popular machine reading comprehension dataset (which we
used for span selection).  Compared to SQuAD, WikiQA and TrecQA are
smaller and we therefore follow \newcite{yang2019simple} and integrate
them for model training.  We show statistics for these datasets in
Table~\ref{tab:qa_stats} and examples in the Appendix.

\paragraph{Implementation Details} 
We used the publicly released BERT
model\footnote{https://github.com/huggingface/pytorch-transformers}
and fine-tuned it on our QA tasks.  For the answer sentence selection
model, BERT was fine-tuned with a learning rate of $3\times 10^{-6}$
and a batch size of 16 for 3 epochs.  For span selection, we adopted a
learning rate of~\mbox{$3\times 10^{-5}$} and a batch size of 64 for 5
epochs.  During inference, the confidence threshold for the relevance
estimator was set to~$\theta = 0.75$ \cite{kratzwald2018adaptive} for
both sentence and passage retrieval.  For the evidence estimator,
$k^{\text{QA}}$~was tuned on the development set.  We obtained~90 and
110~evidence sentences from the sentence selection and span selection
models, respectively. For the centrality estimator, the influence of
the query was set to $\phi=0.15$ \cite{wan2008using,wan2014ctsum}.

We also built an ensemble version of our model, by linearly interpolating
evidence scores from the two estimators based on sentence selection and span
extraction. 
Let $(\mathcal E^{\mathcal S}, q^{\mathcal S})$ and $(\mathcal E^{\mathcal P}, q^{\mathcal P})$ denote the selected sentence sets and their evidence scores produced by the sentence selection estimator and span extraction estimator, respectively. 
We obtain the ensemble score for sentence $e$ via:
\begin{equation}
q_e = \\
\begin{cases} 
	 \mu * q_e^{\mathcal S} + (1-\mu) * q_e^{\mathcal P} & e \in  \mathcal E^{\mathcal S}  \cap \mathcal E^{\mathcal P}\\
	 \mu * q_e^{\mathcal S} & e \in \mathcal E^{\mathcal S} \land e \notin \mathcal E^{\mathcal P}\\
     -\infty & e \notin  \mathcal E^{\mathcal S}
\end{cases}
\end{equation}
where the coefficient was set to $\mu=0.9$.

\begin{table}[t]
\bgroup
\def\arraystretch{1.0}
\centerline{
\begin{small}
\begin{tabular}{@{}l@{~}@{~}c@{~~}c@{~~}cc@{~~}c@{~~}c@{}}  
\hline
\multirow{2}*{\textbf{Systems}} 
& \multicolumn{3}{c}{\textbf{DUC 2006}} 
& \multicolumn{3}{c}{\textbf{DUC 2007}}\\
~ & {R-1} & {R-2} & {R-SU4} & {R-1} & {R-2} & {R-SU4} \\
\hline
\textsc{Gold} & 45.7 &11.2 &  17.0 & 47.9 &14.1 &  19.1  \\
\textsc{Oracle} & 40.6 &\hspace*{.8ex}9.1 &  14.8 & 41.8 &10.4 & 16.0 \\
\textsc{Lead} &  32.1 &\hspace*{.8ex}5.3 & 10.4 & 33.4 & \hspace*{.8ex}6.5 & 11.3 \\
\hline
\multicolumn{7}{c}{Graph-based}\\
\hline
\textsc{LexRank}
& 34.2 & 6.4 & 11.4 & 35.8 & \hspace*{1.2ex}7.7 & 12.7 \\
\textsc{GRSum} 
& \hspace*{.1cm}38.4$^\ast$ & \hspace*{.8ex}7.0$^\ast$ & \hspace*{.1cm}12.8$^\ast$ & 42.0 & 10.3 & 15.6 \\
\textsc{CTSum} 
& ---& --- &  --- & 42.6 & 10.8 & 16.2  \\
\hline
\multicolumn{7}{c}{Autoencoder-based}\\
\hline
\textsc{C-attention} 
& 39.3 & 8.7 &  14.1 & 42.3 &10.7&16.1\\
\textsc{VaeSum} 
&39.6 & 8.9 & 14.3 & 42.1 & 11.0 & 16.4  \\
\hline
\multicolumn{7}{c}{Distantly supervised}\\
\hline
{\sc QuerySum}$_{\mathcal{S}}$
& 41.1 & 9.6 & 15.1
& 42.9 & 11.6 & 16.7 \\
{\sc QuerySum}$_{\mathcal{P}}$ & 41.3 & 9.1 & 15.0 & 43.4 & 11.2 & 16.5 \\
{\sc QuerySum}$_{\mathcal{S}+\mathcal{P}}$  & 41.6 & 9.5 & 15.3 & 43.3 & 11.6 & 16.8 \\
\hline
\end{tabular}
\end{small}
}
\egroup
\caption{\label{tab:auto_eval} 
  System performance on DUC 2006 and 2007. 
  R-1, R-2 and R-SU4 stand for the F1 score of ROUGE~1, 2, and SU4, respectively.  Results with $\ast$ were obtained based on our own implementation.}
\end{table}

\begin{table}[t]
\small
\tabcolsep=0.1cm
\bgroup
\def\arraystretch{1.0}
\centerline{
\begin{tabular}{lr r rr r r}  
\hline
\multirow{2}*{\textbf{Systems}} 
& \multicolumn{3}{c}{\textbf{DUC 2006}} 
& \multicolumn{3}{c}{\textbf{DUC 2007}}\\
~ & {R-1} & {R-2} & {R-SU4} & {R-1} & {R-2} & {R-SU4} \\
\hline
{\sc QuerySum}$_{\mathcal{S}}$ & 41.1 & 9.6 & 15.1  & 42.9 & 11.6 & 16.7 \\
\quad $-$Relevance & $\downarrow$1.1 & $\downarrow$1.0 & $\downarrow$0.8  & $\downarrow$1.5 & $\downarrow$1.4 & $\downarrow$1.2 \\
\quad $-$Evidence &  $\downarrow$1.3 & $\downarrow$0.8 & $\downarrow$0.7 & $\downarrow$0.3 & $\downarrow$0.4 & $\downarrow$0.4\\
\quad $-$Finetuning &  $\downarrow$0.8	&  $\downarrow$0.9	&  $\downarrow$0.8 &  $\downarrow$0.5 & $\downarrow$0.7 & $\downarrow$0.6 \\
\quad $-$Centrality & $\downarrow$2.0 & $\downarrow$1.0 & $\downarrow$0.9 & $\downarrow$2.3 & $\downarrow$1.3 & $\downarrow$1.3 \\
\quad $-\tilde{q}$ distribution& $\downarrow$0.2 & $\downarrow$0.2 & $\downarrow$0.1 & $\uparrow$0.1 & $\downarrow$0.1 & $\downarrow$0.1 \\
\hline
{\sc QuerySum}$_{\mathcal{P}}$  &  41.3 & 9.1 & 15.0 & 43.4 & 11.2 & 16.5 \\
\quad $-$Relevance  &  $\downarrow$0.1 & $\uparrow$0.1 & $\downarrow$0.0 & $\downarrow$0.2 & $\uparrow$0.2 & $\uparrow$0.1  \\
\quad $-$Centrality & $\downarrow$1.5 & $\downarrow$1.4 & $\downarrow$1.2 & $\downarrow$3.2 & $\downarrow$2.1 & $\downarrow$2.0\\
\quad $-\tilde{q}$ distribution& $\downarrow$0.2 & $\downarrow$0.0 & $\downarrow$0.1 & $\downarrow$0.5 & $\downarrow$0.0 & $\downarrow$0.1 \\
\hline
\end{tabular}
}
\egroup
\caption{\label{tab:ablation_study} 
 Results of ablation studies when removing individual modules from our
 framework (absolute performance decrease/increase denoted by $\downarrow$/$\uparrow$). }
\end{table}

\section{Automatic Evaluation}

Following standard practice in DUC evaluations, we used ROUGE as our
automatic evaluation metric\footnote{We used \texttt{pyrouge} with the
  following parameter settings: ROUGE-1.5.5.pl -a -c 95 -m -n 2 -2 4
  -u -p 0.5 -l 250.}  \cite{lin2003automatic} We report F1 for ROUGE-1
(unigram-based), ROUGE-2 (bigram-based), and ROUGE-SU4 (based on skip
bigram with a maximum skip distance of 4).


\paragraph{Model Comparisons}
Our results are summarized in Table~\ref{tab:auto_eval}.  The first
block in the table reports the upper bound performance (\textsc{Gold})
which we estimated by treating a (randomly selected) reference summary
as a hypothetical system output and comparing it against the remaining
(three) ground truth summaries.  \textsc{Oracle} uses reference
summaries as queries to retrieve summary sentences, and
\textsc{{Lead}} returns all leading sentences (up to 250 words) of the
most recent document.

The second block in Table~\ref{tab:auto_eval} compares our model to
various \textit{graph-based} approaches which include:
\textsc{LexRank} \cite{erkan2004lexrank}, a widely used unsupervised
approach based on Markov random walks.
\textsc{LexRank} is query-free; it measures relations between all
sentence pairs in a cluster and sentences recommend other similar
sentences for inclusion in the summary.
\textsc{{GRSum}} \cite{wan2008using}, a Markov random walk model that
integrates query-relevance into a {\sc \bf G}raph {\sc \bf R}anking
algorithm;
and \textsc{{CTSum}} \cite{wan2014ctsum} which is
based on \textsc{GRSum} but 
additionally considers sentence {\sc \bf C}er{\sc \bf T}ainty
information in ranking.

The third group in the table shows the performance of
\textit{autoencoder-based} neural approaches. 
\textsc{{C-attention}} \cite{li2017cascaded} is based on
\textbf{C}ascaded attention with sparsity constraints for compressive
multi-document summarization.  \textsc{{VaeSum}} \cite{li2017salience}
employs a generative model based on \textbf{VA}riational
auto\textbf{E}ncoders \cite{kingma2013auto,rezende2014stochastic} and
a data reconstruction model for sentence salience estimation.
\textsc{VAESum} represents the state-of-the-art of neural systems on
DUC (2006, 2007).\footnote{Similar to our experimental setting, its
  hyperparameters are optimized on a development set. For fair
  comparison, we leave aside a few symbolic approaches that take
  advantage of query expansion techniques, and task-specific
  predictors such as position bias.}  They further integrate their
salience estimation module in an integer linear program which selects
VPs and NPs to create the final summary.

The fourth block in Table~\ref{tab:auto_eval} presents different
variants of our query-focused summarizer which we call {\sc
  QuerySum}. We show automatic results with distant supervision based
on isolated $\mathcal{S}$entences (\textsc{QuerySum}$_{\mathcal{S}}$),
$\mathcal{P}$assages (\textsc{QuerySum}$_{\mathcal{P}}$), and an
ensemble model ({\sc QuerySum}$_{\mathcal{S}+\mathcal{P}}$) which
combines both.  As can be seen, our models outperform strong
comparison systems on both DUC test sets:
\textsc{QuerySum}$_{\mathcal{S}}$ achieves the best R-1 while
\textsc{QuerySum}$_{\mathcal{P}}$ achieves the best \mbox{R-2} and
R-SU4. Perhaps unsurprisingly, both models fall behind the human upper
bound.

\begin{figure}[t]
    \centering   
\begin{tikzpicture}

\definecolor{color4}{rgb}{0.83921568627451,0.152941176470588,0.156862745098039}
\definecolor{color0}{rgb}{1,0.498039215686275,0.0549019607843137}
\definecolor{color3}{rgb}{0.580392156862745,0.403921568627451,0.741176470588235}
\definecolor{color2}{rgb}{0.172549019607843,0.627450980392157,0.172549019607843}
\definecolor{color1}{rgb}{0.12156862745098,0.466666666666667,0.705882352941177}
\definecolor{color5}{rgb}{0.549019607843137,0.337254901960784,0.294117647058824}

\begin{axis}[
width=0.5\textwidth,
height=0.35\textwidth, 
legend cell align={left},
legend columns=2,
legend style={
nodes={scale=0.8, transform shape},
at={(0.97,0.03)}, anchor=south east, draw=white!80.0!black
},
x grid style={lightgray!92.02614379084967!black},
xlabel={\(\scriptstyle k^\mathrm{{IR}}\)},
xmin=0, xmax=510,
y grid style={lightgray!92.02614379084967!black},
ylabel={\small ROUGE-2 Recall},
ymin=0, ymax=60,
ticklabel style={font=\small}
]
\addplot [color0, mark=+, mark size=2, mark options={solid}]
table {%
10 5.98
20 9.96
30 13.06
40 15.81
50 17.86
60 19.73
70 21.73
80 23.41
90 24.9
100 26.09
110 27.17
120 28.41
130 29.37
140 30.32
150 31.33
160 32.24
170 32.93
180 33.62
190 34.25
200 34.96
210 35.46
220 35.95
230 36.46
240 36.93
250 37.53
260 37.92
270 38.26
280 38.54
290 38.8
300 39.02
310 39.33
320 39.58
330 39.74
340 39.88
350 40.02
360 40.13
370 40.25
380 40.32
390 40.38
400 40.51
410 40.6
420 40.69
430 40.74
440 40.8
450 40.94
460 40.97
470 40.99
480 41.01
490 41.03
500 41.05
};
\addlegendentry{Full$_{\mathcal S}$}
\addplot [color3, mark=+, mark size=2, mark options={solid}]
table {%
10 18.67
20 26.09
30 31.49
40 35.47
50 38.47
60 40.82
70 42.63
80 44.45
90 45.7
100 46.82
110 47.73
120 48.48
130 49.17
140 49.73
150 50.19
160 50.5
170 50.79
180 51.05
190 51.23
200 51.43
210 51.61
220 51.8
230 51.86
240 51.96
250 52.02
260 52.07
270 52.1
280 52.12
290 52.14
300 52.16
310 52.19
320 52.2
330 52.21
340 52.21
350 52.21
360 52.22
370 52.22
380 52.22
390 52.22
400 52.22
410 52.22
420 52.22
430 52.22
440 52.22
450 52.22
460 52.22
470 52.22
480 52.22
490 52.22
500 52.22
};
\addlegendentry{Full$_{\mathcal P}$}
\addplot [color1]
table {%
10 5.78
20 9.47
30 12.57
40 14.95
50 17.13
60 19.23
70 21.03
80 22.87
90 24.37
100 25.54
110 26.76
120 27.76
130 28.7
140 29.72
150 30.73
160 31.45
170 32.12
180 32.91
190 33.58
200 34.21
210 34.73
220 35.3
230 35.77
240 36.1
250 36.54
260 36.94
270 37.28
280 37.65
290 37.87
300 38.07
310 38.32
320 38.56
330 38.77
340 38.9
350 39.06
360 39.14
370 39.27
380 39.41
390 39.48
400 39.56
410 39.61
420 39.68
430 39.73
440 39.76
450 39.83
460 39.85
470 39.92
480 39.93
490 39.96
500 39.98
};
\addlegendentry{Narrative$_{\mathcal S}$}
\addplot [color4]
table {%
10 18.38
20 26.43
30 31.39
40 35.07
50 38.15
60 40.47
70 42.45
80 44.12
90 45.54
100 46.66
110 47.53
120 48.22
130 49
140 49.47
150 49.87
160 50.21
170 50.59
180 50.96
190 51.16
200 51.33
210 51.47
220 51.61
230 51.68
240 51.76
250 51.83
260 51.86
270 51.9
280 51.93
290 51.97
300 51.98
310 52.02
320 52.02
330 52.02
340 52.03
350 52.03
360 52.03
370 52.03
380 52.03
390 52.03
400 52.03
410 52.03
420 52.03
430 52.03
440 52.03
450 52.03
460 52.03
470 52.03
480 52.03
490 52.03
500 52.03
};
\addlegendentry{Narrative$_{\mathcal P}$}
\addplot [color2, mark=x, mark size=2, mark options={solid}]
table {%
10 5.74
20 9.65
30 12.67
40 15.3
50 17.16
60 18.98
70 20.69
80 21.95
90 23.16
100 24.28
110 25.4
120 26.36
130 27.29
140 28.07
150 28.68
160 29.41
170 30.04
180 30.58
190 31.17
200 31.59
210 31.98
220 32.23
230 32.56
240 32.84
250 32.99
260 33.12
270 33.22
280 33.26
290 33.33
300 33.35
310 33.44
320 33.48
330 33.51
340 33.56
350 33.6
360 33.64
370 33.67
380 33.71
390 33.75
400 33.78
410 33.82
420 33.93
430 33.98
440 34.03
450 34.06
460 34.07
470 34.09
480 34.1
490 34.13
500 34.14
};
\addlegendentry{Title$_{\mathcal S}$}
\addplot [color5, mark=x, mark size=2, mark options={solid}]
table {%
10 18.27
20 25.41
30 30.84
40 34.53
50 37.77
60 39.91
70 41.76
80 43.27
90 44.45
100 45.25
110 46.04
120 46.77
130 47.32
140 47.75
150 48.03
160 48.24
170 48.41
180 48.56
190 48.69
200 48.76
210 48.87
220 48.95
230 49.01
240 49.05
250 49.09
260 49.11
270 49.11
280 49.12
290 49.14
300 49.17
310 49.17
320 49.18
330 49.18
340 49.19
350 49.19
360 49.2
370 49.2
380 49.2
390 49.2
400 49.2
410 49.2
420 49.2
430 49.2
440 49.2
450 49.2
460 49.2
470 49.2
480 49.2
490 49.2
500 49.2
};
\addlegendentry{Title$_{\mathcal P}$}
\end{axis}

\end{tikzpicture}
   \vspace{-1.5em}
   \caption{\label{fig:ir_curve} Performance (ROUGE-2 Recall)
     over~$k^{\text{IR}}$ best retrieved segments (development set).
	${\mathcal S}$ and ${\mathcal P}$ refer to sentence and passage retrieval, respectively.
     \textit{Full} is the concatenation of the query title and
     narrative.}
\end{figure}
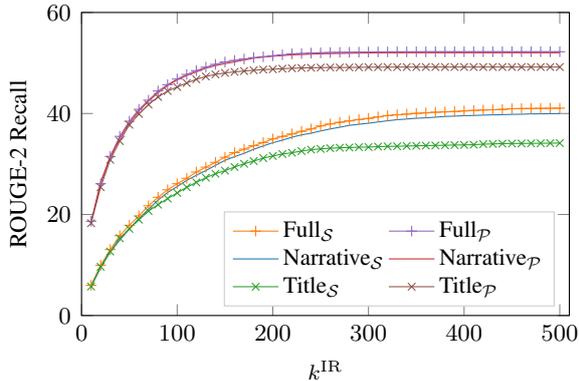

\begin{figure}[t]
    \centering
\begin{tikzpicture}

\definecolor{color0}{rgb}{1,0.498039215686275,0.0549019607843137}
\definecolor{color3}{rgb}{0.580392156862745,0.403921568627451,0.741176470588235}
\definecolor{color2}{rgb}{0.172549019607843,0.627450980392157,0.172549019607843}
\definecolor{color1}{rgb}{0.12156862745098,0.466666666666667,0.705882352941177}
\definecolor{color5}{rgb}{0.549019607843137,0.337254901960784,0.294117647058824}
\definecolor{color4}{rgb}{0.83921568627451,0.152941176470588,0.156862745098039}

\begin{axis}[
width=0.5\textwidth,
height=0.3\textwidth,       
legend cell align={left},
legend style={
nodes={scale=0.8, transform shape},
at={(0.97,0.03)}, anchor=south east, draw=white!80.0!black},
tick align=inside,
tick pos=both,
x grid style={lightgray!92.02614379084967!black},
xlabel={\(\scriptstyle k^\mathrm{{QA}}\)},
xmin=0, xmax=205,
y grid style={lightgray!92.02614379084967!black},
ylabel={\small ROUGE-2 Recall},
ymin=5, ymax=40,
ticklabel style={font=\small}
]
\addplot [color0, mark=x, mark size=2, mark options={solid}]
table {%
10 6.96
20 10.99
30 14.1
40 16.98
50 19.51
60 21.43
70 23.01
80 24.53
90 26.05
100 27.31
110 28.48
120 29.52
130 30.77
140 31.89
150 32.78
160 33.58
170 34.38
180 35.1
190 35.8
200 36.32
};
\addlegendentry{Span extraction}
\addplot [color1, mark= +, mark size=2, mark options={solid}]
table {%
10 7.61
20 12.2
30 15.53
40 18.15
50 20.27
60 21.97
70 23.42
80 24.64
90 25.9
100 26.71
110 27.45
120 28.09
130 28.73
140 29.24
150 29.66
160 29.95
170 30.23
180 30.44
190 30.6
200 30.73
};
\addlegendentry{Sentence selection}
\addplot [color2]
table {%
10 5.78
20 9.47
30 12.57
40 14.95
50 17.13
60 19.23
70 21.03
80 22.87
90 24.37
100 25.54
110 26.76
120 27.76
130 28.7
140 29.72
150 30.73
160 31.45
170 32.12
180 32.91
190 33.58
200 34.21
};
\addlegendentry{Sentence retrieval}
\end{axis}

\end{tikzpicture}
     \vspace{-1.5em}
     \caption{\label{fig:qa_curve} Performance  (ROUGE-2
       Recall) over~$k^{\text{QA}}$ best evidence sentences selected
       by estimators trained on sentences and passages (development set).}
\end{figure}
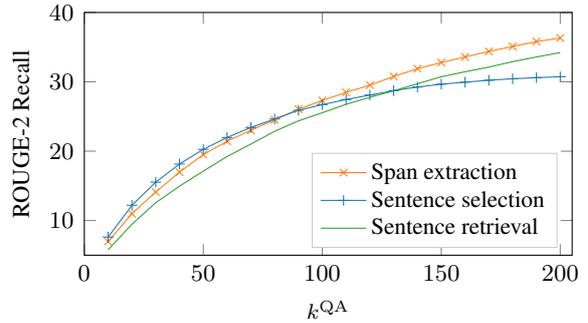

\paragraph{Ablation Studies}
We also conducted ablation experiments to verify the effectiveness of
the proposed coarse-to-fine framework.  We present results in
Table~\ref{tab:ablation_study} for \textsc{QuerySum}$_{\mathcal{S}}$
and \textsc{QuerySum}$_{\mathcal{P}}$ when individual modules are
removed. In the \mbox{$-$Relevance} setting, all text segments
(i.e.,~sentences or passages) in a cluster are given as input to the
evidence estimator module. The \mbox{$-$Evidence} setting treats all
retrieved segments as evidence for summarization. Note that since our
summarizer operates on sentences, we can only assess this
configuration with the \textsc{QuerySum}$_{\mathcal{S}}$ model; we
take the top $k^{\text{QA}}$ sentences from the retrieval module as
evidence.
The \mbox{$-$Centrality} setting treats the (ranked) output of the
evidence estimator as the final summary.

As can be seen, removing the retrieval module leads to a large drop in
the performance of \textsc{QuerySum}$_{\mathcal{S}}$.  This indicates
that the (deep) semantic matching model trained for sentence selection
can get distracted by noise which a (shallow) relevance matching model
can help pre-filter.  Interestingly, when the matching model is
trained on passages, the retrieval module seems more or less
redundant, there is in fact a slight improvement in ROUGE scores,
except for R-1 (see row \textsc{QuerySum}$_{\mathcal{P}}$, $-$
Relevance in Table~\ref{tab:ablation_study}). This suggests that the
evidence estimator trained on passages is more robust and captures the
semantics of the query more faithfully. Moreover, since it takes
contextual signals into account, it is able to recognize irrelevant
information and unanswerability is explicitly modeled.  We show in
Figure~\ref{fig:ir_curve} how ROUGE-2 varies over~$k^{\text{IR}}$ best
retrieved segments.
We compare three different types of query
settings, the short \textit{title}, the \textit{narrative}, and the
\textit{full} query with both the title and the narrative.  As
expected, recall increases with $k^{\text{IR}}$ (i.e., when more
evidence is selected) and then finally converges. For both sentence and passage retrieval settings,
the full query achieves best performance
over $k^{\text{IR}}$, with the narrative being most informative when
it comes to relevance estimation.

Performance also drops in Table~\ref{tab:ablation_study} when the
evidence estimator is removed (see \textsc{QuerySum}$_{\mathcal{S}}$,
\mbox{$-$Evidence} in Table~\ref{tab:ablation_study}).  In
Figure~\ref{fig:qa_curve}, we plot how ROUGE-2 varies with
increasing~$k^{\text{QA}}$ when the evidence component is estimated on
passages and sentences for the full model. As can be seen, the model
trained on passages surpasses the model trained on sentences roughly
when $k^{\text{QA}}=80$.  For comparison, we also show the performance
of the retrieval module by treating the top sentences as evidence.
The retrieval curve is consistently under the passage curve, and under
the sentence curve when~$k^{\text{QA}}<140$.  Since the quality of top
sentences directly affects the quality of the summarization module,
this further demonstrates the effectiveness of evidence estimation in
terms of reranking retrieved segments.

Finally, Table~\ref{tab:ablation_study} shows that the removal of the
centrality estimator  decreases performance even when the query
and appropriate evidence are taken into account. These experiments
suggest that the centrality estimator further learns to select
important sentences from the available evidence that are summary
worthy. We further assessed the effectiveness of the query
component~$\tilde{q}$ in estimating centrality (see
Equation~\eqref{eq:query}) by setting $\tilde{q}$ to a uniform
distribution. Again, we observe that in most cases performance
decreases which indicates that this component provides a slight
benefit over and above filtering segments according to their relevance
to the query. 

\begin{table}[t]
\small
\centering
\begin{tabular}{lllll} \hline
  \textbf{Method} & \textbf{Rel} & \textbf{Suc} & \textbf{Coh}  & \textbf{All}
  \\\hline
  {\sc Lead}		& 3.75$^{\triangleright\dagger\circ}$ & 3.60$^{\dagger\circ}$ &	4.27$^{\triangleright}$ & 3.96$^{\dagger\circ}$\\
  {\sc VAESum} &  4.28	& 3.62$^{\dagger\circ}$	& 4.05$^{\dagger\circ}$ & 4.03$^{\dagger\circ}$\\
{\sc QuerySum} & 4.32	& 3.93$^{\triangleright}$	& 4.27$^{\triangleright}$ & 4.22$^{\triangleright}$ \\ \hline
{\sc Gold}		& 4.36  & 3.93$^{\triangleright}$& 4.35$^{\triangleright}$ & 4.26$^{\triangleright}$ \\

\hline
\end{tabular}
\caption{\label{tab:summ_results} Human evaluation results: average
  \textbf{Rel}evance, \textbf{Suc}cinctness, \textbf{Coh}erence ratings;
  \textbf{All} is the average across ratings;  $\triangleright$: sig different
  from \textsc{VAESum}; $\dagger$:   sig different from
  \textsc{QuerySum}; $^\circ$: sig different
  from Gold
  (at \mbox{$p < 0.1$},  using a pairwise t-test). 
}
\end{table}

\section{Human Evaluation}
\label{sec:human-evaluation}

We further evaluated the summaries created by our model in a judgment
elicitation study via Amazon Mechanical Turk. Specifically, native
English speakers (self-reported) were provided with query-summary
pairs and asked to rate the summaries on two dimensions:
\textit{Succinctness} (does the summary avoid unnecessary detail and
redundant information?) and \textit{Coherence} (does the summary make
logical sense?). The ratings were obtained using a five point Likert
scale. In addition, participants were asked to assess the
\emph{Relevance} of the summary to the query. Specifically,
crowdworkers read a summary and for each sentence decided whether it
is relevant (i.e.,~whether it provides an answer to the query),
irrelevant (i.e., it does not answer the query), and partially
relevant (i.g., there is some related information but it is not clear
it directly answers the query). Relevant sentences were awarded a
score of~5, partially relevant ones were given a score of~2.5, and
0~otherwise. Sentence scores were averaged to obtain a relevance score
for the whole summary.

Participants assessed summaries created by \textsc{VAESum}\footnote{We
  are grateful to Piji Li for providing us with the output of their system.},
the previous state-of-the-art system, \textsc{QuerySum}, and the
\textsc{Lead} baseline. We also included a randomly selected
\textsc{Gold} standard summary as an upper bound.  We sampled 20
clusters from DUC 2006 and 2007 test sets (10 from each set) and
collected three responses per query-summary pair.
Table~\ref{tab:summ_results} shows the ratings for each system.  As
can be seen, participants find {\sc QuerySum} summaries more relevant
to the queries and with less redundant information compared to
\textsc{Lead} and \textsc{VAESum}.  Our multi-step estimation process
also produces more coherent summaries (as coherent as {\sc Lead}) even
though coherence is not explicitly modeled.  
Overall, participants perceive \textsc{QuerySum} summaries as
significantly better ($p<0.1$) compared to \textsc{Lead} and
\textsc{VAESum} (see the Appendix for examples of system output). 
\section{Conclusions}
\label{sec:conclusions}

In this work, we proposed a coarse-to-fine estimation framework for
query focused multi-document summarization.  We explored the potential
of leveraging distant supervision signals from Question Answering to
better capture the semantic relations between queries and document
segments.  Experimental results across datasets show that the proposed
model yields results superior to competitive baselines contributing to
summaries which are more relevant and less redundant. We have also
shown that disentangling the tasks of relevance, evidence, and
centrality estimation is beneficial allowing us to progressively
specialize the summaries to the semantics of the query. In the future,
we would like to generate abstractive summaries following an
unsupervised approach \cite{baziotis-etal-2019-seq,chu2019meansum} and
investigate how recent advances in open domain QA
\cite{wang2019multi,qi2019answering} can be adapted for query focused
summarization.

\bibliography{acl2020.bib}
\bibliographystyle{acl_natbib}
\end{document}